\documentclass[10pt,twocolumn,letterpaper]{article}

\usepackage[pagenumbers]{cvpr} % To force page numbers, e.g. for an arXiv version

\definecolor{cvprblue}{rgb}{0.21,0.49,0.74}
\usepackage[pagebackref,breaklinks,colorlinks,allcolors=cvprblue]{hyperref}

\def\paperID{28339} % *** Enter the Paper ID here
\def\confName{CVPR}
\def\confYear{2026}

\title{High-Fidelity Diffusion Face Swapping with ID-Constrained Facial Conditioning}

\author{%
Dailan He$^{1,2}$~~~ Xiahong Wang$^{2}$~~~ 
Shulun Wang$^{2}$~~~ Guanglu Song$^{2}$\\
Bingqi Ma$^{2}$~~~ Hao Shao$^{1}$~~~ 
Yu Liu$^{2}$ ~~~ Hongsheng Li$^{1,3,4}$ \\ 
\\
$^1$CUHK MMLab ~~~ $^2$Vivix Group Limited ~~~ $^3$Shenzhen Loop Area Institute ~~~ $^4$CPII under InnoHK \\
{\tt\small hedailan@link.cuhk.edu.hk ~~~~ hsli@ee.cuhk.edu.hk}
}

\begin{document}

% \maketitle

%%% https://github.com/cvpr-org/author-kit/discussions/27
\twocolumn[{%
\renewcommand\twocolumn[1][]{#1}%
\maketitle
\centering
\includegraphics[width=0.9981\linewidth]{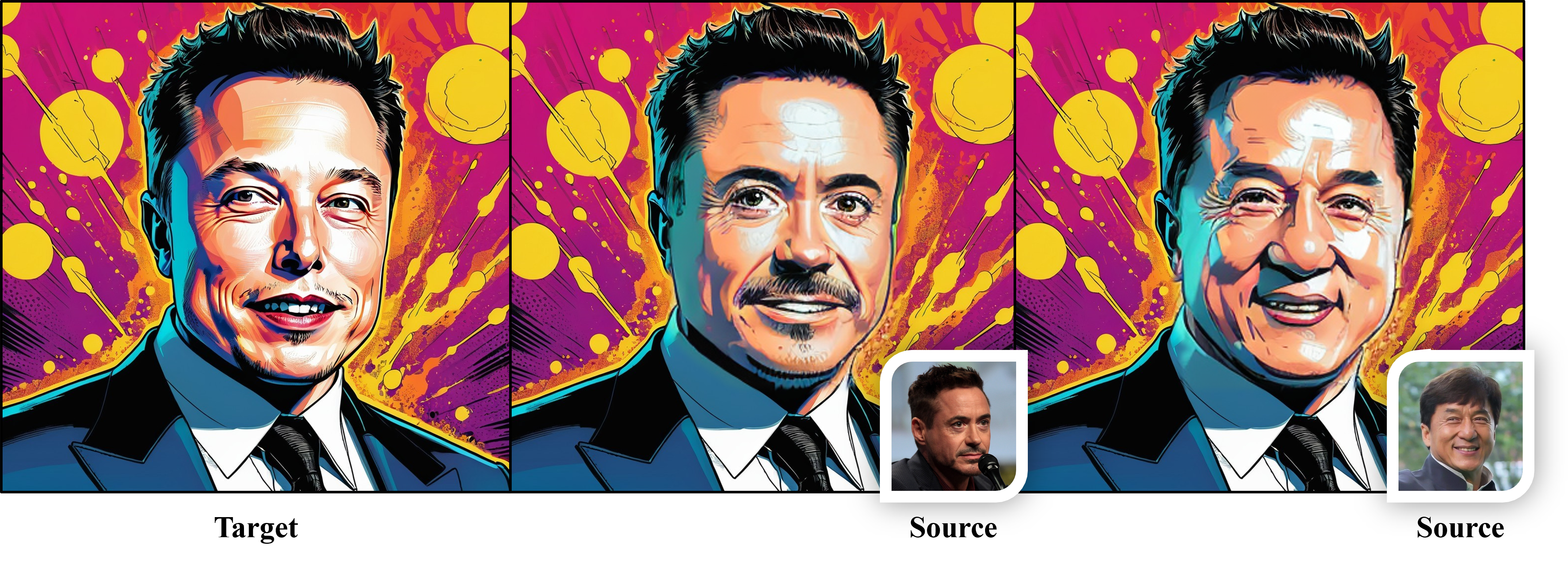}
\captionof{figure}{Face-swapping examples of proposed approach. Given the \textit{source} face (white-border) and \textit{target} face (left), the face-swapping applications aim to transfer the identity of the source image to the attributes (expression, pose, \etc) of the target face. \vspace{1em}}
\label{fig:teaser}
}]

\begin{abstract}
Face swapping aims to seamlessly transfer a source facial identity onto a target while preserving target attributes such as pose and expression. Diffusion models, known for their superior generative capabilities, have recently shown promise in advancing face-swapping quality. 
This paper addresses two key challenges in diffusion-based face swapping: the prioritized preservation of identity over target attributes and the inherent conflict between identity and attribute conditioning. To tackle these issues, we introduce an identity-constrained attribute-tuning framework for face swapping that first ensures identity preservation and then fine-tunes for attribute alignment, achieved through a decoupled condition injection. We further enhance fidelity by incorporating identity and adversarial losses in a post-training refinement stage. Our proposed identity-constrained diffusion-based face-swapping model outperforms existing methods in both qualitative and quantitative evaluations, demonstrating superior identity similarity and attribute consistency, achieving a new state-of-the-art performance in high-fidelity face swapping.

\end{abstract}  
\section{Introduction}
\label{sec:intro}

Face swapping~\cite{deepfakes,perov2020deepfacelab,chen2020simswap,baliah2024reface,kim2022diffface,zhao2023diffswap,gao2021infoswap,thies2016face2face,li2019faceshifter} aims to transfer a source face onto a target face while preserving the target attributes, such as pose and expression (Figure~\ref{fig:teaser}). This technology has significant applications in fields like film production, gaming, and digital twins, making it an essential area of research in human character generation. Traditional face-swapping methods often rely on generative adversarial networks (GANs)~\cite{nirkin2019fsgan,chen2020simswap,goodfellow2014gan}. Those methods often suffer from issues like limited image quality and mode collapse, due to the limitations of GAN itself. Intuitively, it is possible to promote face fidelity by using more powerful generative models, such as diffusion models~\cite{ho2020ddpm,rombach2022latentdiff}. 
The advent of diffusion models~\cite{ho2020ddpm,song2019scorematching,rombach2022latentdiff,song2020ncsnpp} has marked a substantial improvement in data generation, advancing sample quality beyond that of most previous models in both unconditional~\cite{dhariwal2021adm,ho2020ddpm,song2019scorematching,song2020ncsnpp} and conditional~\cite{rombach2022latentdiff, ye2023ipadapter, li2023gligen, ruiz2023dreambooth,zhang2023controlnet,zong2024easyref,ma2024exploring} generation tasks. Recent studies~\cite{kim2022diffface,zhao2023diffswap,baliah2024reface,han2025faceadapter} also demonstrate competitive results in face swapping when leveraging diffusion-based approaches, pushing the boundaries of what is possible in face-swapping technology. To further enhance this line of approaches, we delve into the two essential facial condition inputs: identity and attribute.
Though disentangling and injecting those facial features has been widely discussed in GAN-based face swapping~\cite{chen2020simswap,gao2021infoswap,liu2023e4s,wang2021hififace}, it is under investigation in diffusion approaches, to the best of our knowledge. Due to the principal difference between the two generative models, a new design for the diffusion-based framework will be needed.

\begin{figure}
    \centering
    \includegraphics[width=0.8\linewidth]{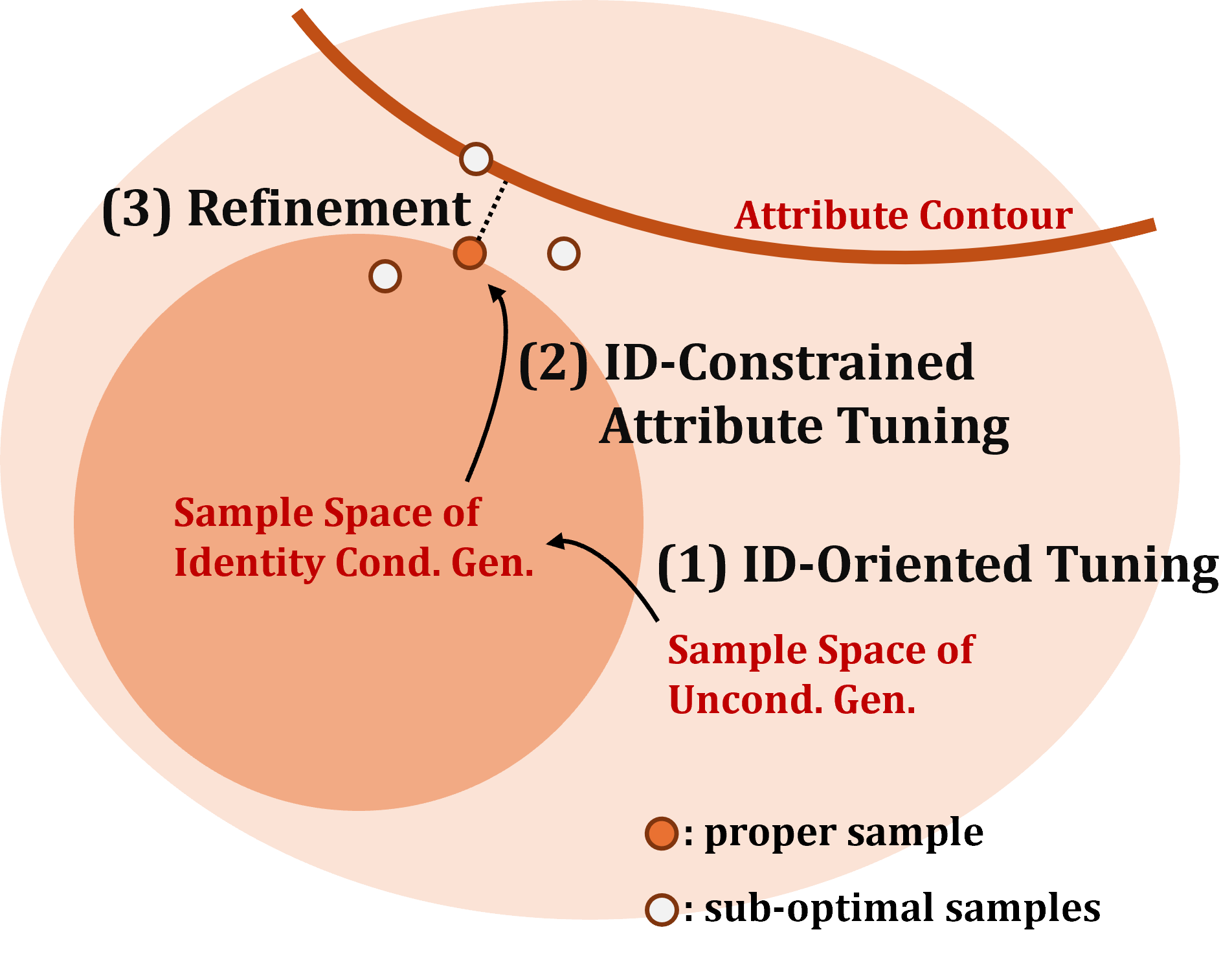}
        \caption{The identity-constrained attribute tuning scheme. We aim to obtain the red-dot sample with constrained identity and attribute. We propose to divide this process into 3 stages.}
    \label{fig:solution-region}
\end{figure}

In this paper, we follow an intuition about those conditions: preserving the source identity is prioritized over maintaining target attributes in face swapping. We \textit{primarily} expect the swapping results to ``seem like" the source face identity enough, and then \textit{secondarily} require the result to follow the expression and poses of the given target. In other words, we expect an \textit{identity-constrained} solution, presented by red dot in Figure~\ref{fig:solution-region}.
We propose to achieve the objective in a multi-stage manner, which we name \textbf{the identity-constrained facial conditioning scheme}. We first fine-tune an unconditional diffusion model using only identity conditions, thereby narrowing the model's solution set to identity-consistent outputs (orange region in Figure~\ref{fig:solution-region}). After constraining the output identity, we inject the attribute conditions and guide the model toward capturing the target expression and poses, balancing identity and attribute alignment. 
To further enhance identity fidelity, we apply an end-to-end refinement process using identity and adversarial losses~\cite{goodfellow2014gan}. 
To mitigate the interference of training with the two types of conditions, we propose \textbf{the decoupled identity and attribute condition injection}, with data-wise condition decoupling and deliberately designed identity and attribute representation.
To avoid overfitting and condition leakage, we propose to use paired images from the same identity with different attributes as training data. Then we carefully extract the entirely decoupled identity and attribute features from two independent calculation paths with the two different inputs. To encode the identity, we use the pretrained face recognition model
as well as DINOv2~\cite{oquab2023dinov2} on the source face image. To obtain the attribute features, we feed the target face to the feature encoder of SimSwap~\cite{chen2020simswap}, which is successfully adopted in GAN-based face swapping.
Hereby, our model trained on all the proposed techniques achieves state-of-the-art qualitative and quantitative face-swapping performance.

\begin{figure}[t]
    \centering
        \includegraphics[width=0.78\linewidth]{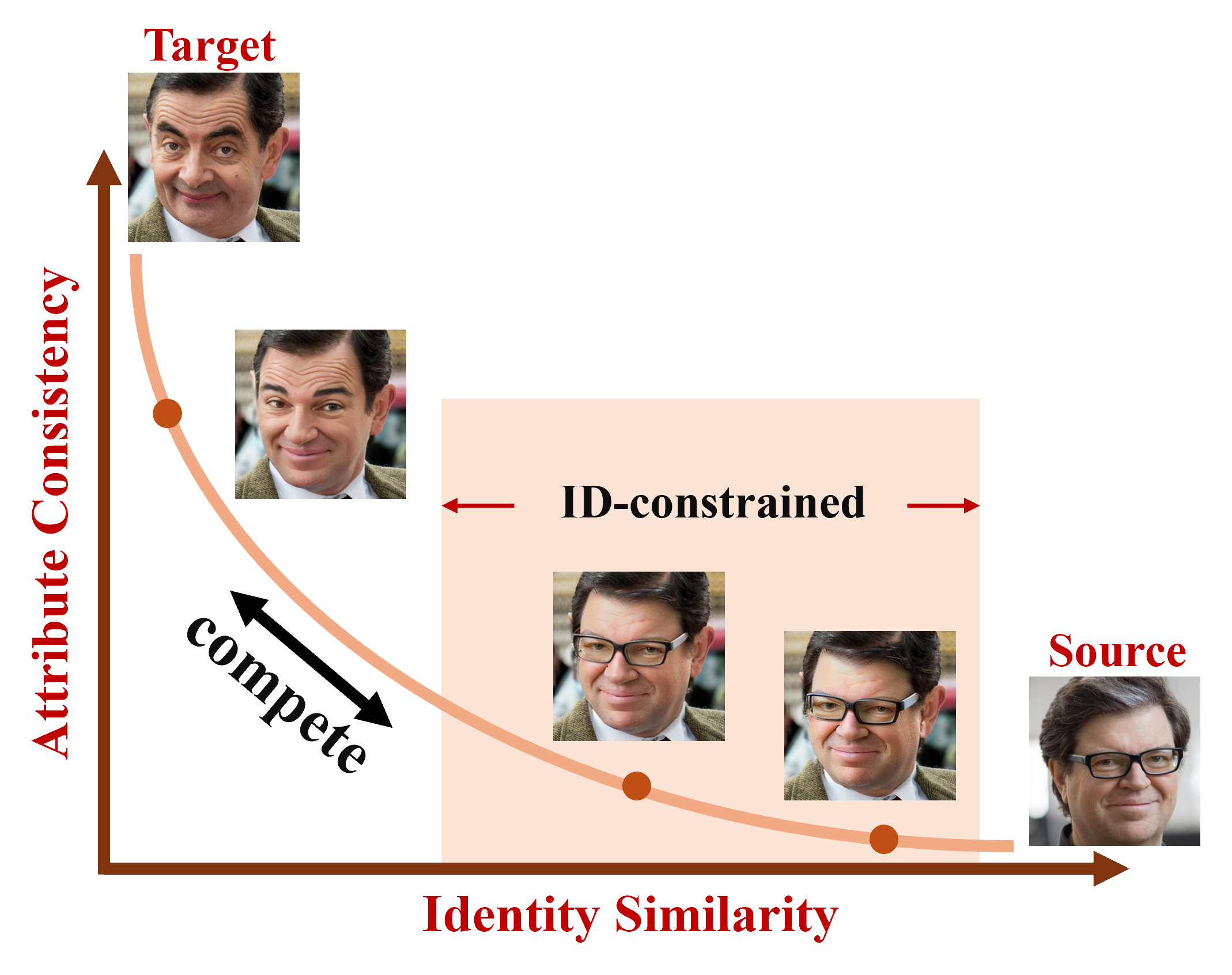}
    \caption{Identity and attribute compete against each other. The former pushes the model output closer to the source face, while the latter aligns the model to the target face.}
    \label{fig:insight}
\end{figure}

In contrast, existing approaches often jointly feed all the conditions in a single-stage training scheme. We found that identity and attribute conditioning often work in opposition during training (See Figure~\ref{fig:insight}). While identity preservation pushes the model to output a face closer to the source face, attribute consistency encourages alignment with the target face. It implies that training a diffusion model jointly conditioned on both identity and attribute might be challenging as the two conditions might require opposite adaptations. 
On the other hand, an attribute-constrained scheme 
might struggle with face shape mismatches, leading to distortion.

In this paper, we make several contributions to the diffusion-based face-swapping approach, which we summarize as follow:
\begin{enumerate}
    \item We propose the decoupled identity and attribute condition injection, to address challenges in joint injection and enable the multi-stage training.
    \item We propose the identity-constrained facial conditioning scheme, to guide the model fit identity and attribute conditioning in a multi-step manner.
    \item We present a novel diffusion-based model with state-of-the-art face-swapping performance.
\end{enumerate}

\section{Related works}
\label{sec:related-works}

\subsection{Diffusion models}

Diffusion models have emerged as a powerful framework for generative modeling, with significant advancements in recent years. One of the pioneering approaches~\cite{song2019scorematching,song2020ncsnpp,ho2020ddpm}, the Denoising Diffusion Probabilistic Model (DDPM)~\cite{ho2020ddpm}, introduced a generative process that learns to reverse a multi-step noise perturbation of data.  By training a neural network to iteratively denoise a noisy signal, DDPM achieves high-quality sample generation, and its probabilistic formulation allows for theoretical guarantees of sample fidelity.
Latent Diffusion Models (LDMs)~\cite{rombach2022latentdiff} introduce a crucial optimization by applying the diffusion process in a lower-dimensional latent space, rather than in the original high-dimensional pixel space. This compression reduces computational costs while preserving the quality of generated images. Its text-to-image conditional adapts such as Stable Diffusion have become standard foundation generative models and are widely investigated in the community. Denoising Diffusion Implicit Model (DDIM)~\cite{song2020ddim} builds upon DDPM and is also available on LDMs, offering a more efficient generation process by skipping some denoising steps through a non-Markovian sampling process. This approach maintains the sample quality while significantly reducing the number of sampling steps required for generation, enabling faster image generation.

\subsection{Face swapping}

There are two sorts of modern face-swapping methods: optimization-based and end-to-end methods. The former such as Deepfakes~\cite{deepfakes} requires specific optimization for each identity input, which might achieve competitive results but is extremely time-consuming when a large-scale batched operation is required. On the other hand, data-driven end-to-end approaches~\cite{thies2016face2face,nirkin2018face,li2019faceshifter,perov2020deepfacelab,wang2021hififace,zhang2023towards,ma2022rethinking} attract plenty of attention because of their training-free manner, mentioned as subject-agnostic property~\cite{nirkin2019fsgan} in literature. Among them,
FSGAN~\cite{nirkin2019fsgan} proposes an early implementation of end-to-end face swapping, with a reenactment network and subject-agnostic design.
SimSwap~\cite{chen2020simswap} achieves a good id-attribute preservation balance by adopting explicit identity injection and adapted GAN training.
InfoSwap~\cite{gao2021infoswap} investigates the information bottleneck to disentangle identity from representation and optimize the generated identity to a proper distance from the target identity.
E4S~\cite{liu2023e4s} alternatively adopts the GAN inversion to figure out disentangled facial components of the source and target and remix them for face swapping. Most of those end-to-end approaches depend on GAN, which ensures superior generation ability and high-fidelity face-swapping results. Recently, investigations on face-swapping approaches are rapidly turning to diffusion models, because of their even stronger generation ability~\cite{dhariwal2021adm}. 
DiffFace~\cite{kim2022diffface} performs early attempts to swap faces with diffusion inpainting models. It proposes to enhance the gradient-based facial generation which is time-consuming.
DiffSwap~\cite{zhao2023diffswap} 
employs 3D Morphable Model (3DMM)~\cite{blanz20233dmm} to pre-swap the given source and target as a prior condition. It also proposes to adopt ID loss on mid-point prediction to enhance identity similarity. 
REFace~\cite{baliah2024reface} proposes to inject multiple visual features and train the diffusion model accompanied by ID supervision on a few-step DDIM sampling, achieving competitive face-swapping performance and faster inference speed than DiffSwap and DiffFace.
FaceAdapter~\cite{han2025faceadapter} proposes to train an adapter-like facial generation plugin to fit face swapping on various text-to-image diffusion models.

\section{Methodology}

To utilize the strengths of pre-trained diffusion models, we aim to inject facial conditions into a general-purpose conditional diffusion model and fine-tune it for face swapping. This approach can be viewed as training an image inpainting model that generates a facial foreground conditioned on source identity and target attributes while maintaining the target background~\cite{baliah2024reface,zhao2023diffswap}.

In this section, we introduce our proposed identity-constrained diffusion training framework for face swapping. First, we will revisit the foundational concepts and notation of diffusion models. Subsequently, we will discuss the decoupled facial condition injection and identity-constrained facial conditioning in sequence.

\subsection{Preliminary: diffusion models}
\label{sec:preliminary-of-diff-model}

Let $\mathcal{N}(x_{t}|x_{t-1}): x_0\to x_T$ denote a $T$-step Markov Gaussian process from the original image $x_0$ to pure noise $x_T$. The objective of diffusion model training is to approximate its step-wise posterior, $P_\theta(x_{t-1}|x_t)$.
Since the forward process involves adding noise, the reverse sampling process $P_\theta: x_T \to x_0$ becomes a denoising process, gradually recovering an image $x_0$ from pure noise $x_T$. Intuitively, image generation using a diffusion model can be viewed as a multi-step denoising process, where a UNet predicts the noise in the noisy sample $x_t$ at each time step $t$ to obtain a progressively clearer image, $x_{t-1}$. Training this denoising process involves minimizing the KL-divergence between the true posterior, $\mathcal{N}(x_{t-1}|x_t, x_0)$, and the parametric approximation, $P_\theta(x_{t-1}|x_t)$, resulting in a diffusion loss in MSE form~\cite{ho2020ddpm,kingma2021vdm}:
\begin{align}
    \mathcal{L}_\mathrm{diff} &= \sum_{t=1}^T D_{\mathrm{KL}}[\mathcal{N}(x_{t-1}|x_t, x_0)\|P_\theta(x_{t-1}|x_t)]
    \label{eq:ori-kld-diff-loss}
    \\
    &= \sum_{t=1}^T \lambda_t \|\epsilon_\theta(x_t; t) - \epsilon \|^2_2,
\end{align}
where $\epsilon_\theta(\cdot)$ is the noise-prediction model with learnable parameters $\theta$. It is often a UNet~\cite{ho2020ddpm,song2019scorematching}. $\epsilon$ is the noise disturbing the image. The scale factor $\lambda_t$ is a weight variable about the time steps $t$ and is usually omitted (\ie set to constant ones) empirically to obtain better sample quality~\cite{ho2020ddpm,kingma2021vdm}. To train this loss function, a common practice is to adopt Monte Carlo sampling along time steps $t$.  

For text-to-image generation, a common approach is to use a pre-trained latent diffusion model~\cite{rombach2022latentdiff}, applying the denoising generation process in a latent space defined by a standard variational autoencoder (VAE)~\cite{dkingma2014vae,rombach2022latentdiff}. The text tokens as conditions are fed into the cross-attention layers of denoising UNets.

\begin{figure}[t]
    \centering
    \begin{subfigure}{0.9\linewidth}
    \centering
    \includegraphics[width=0.8\linewidth]{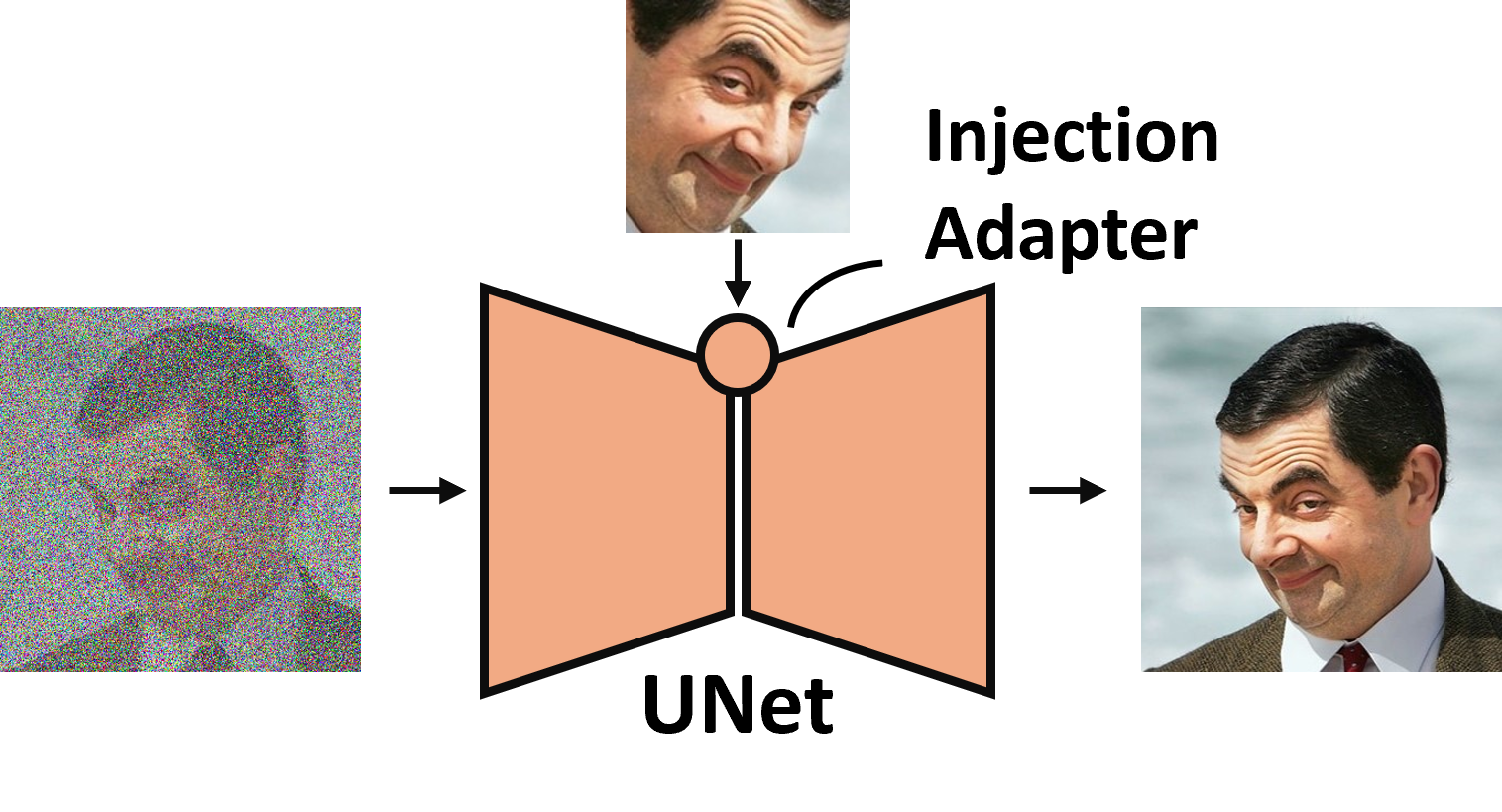}
    \caption{Augmentation-based condition injection.}
    \label{fig:aug-injection}
    
    \end{subfigure}

    \begin{subfigure}{0.9\linewidth}
    \centering
    \includegraphics[width=0.8\linewidth]{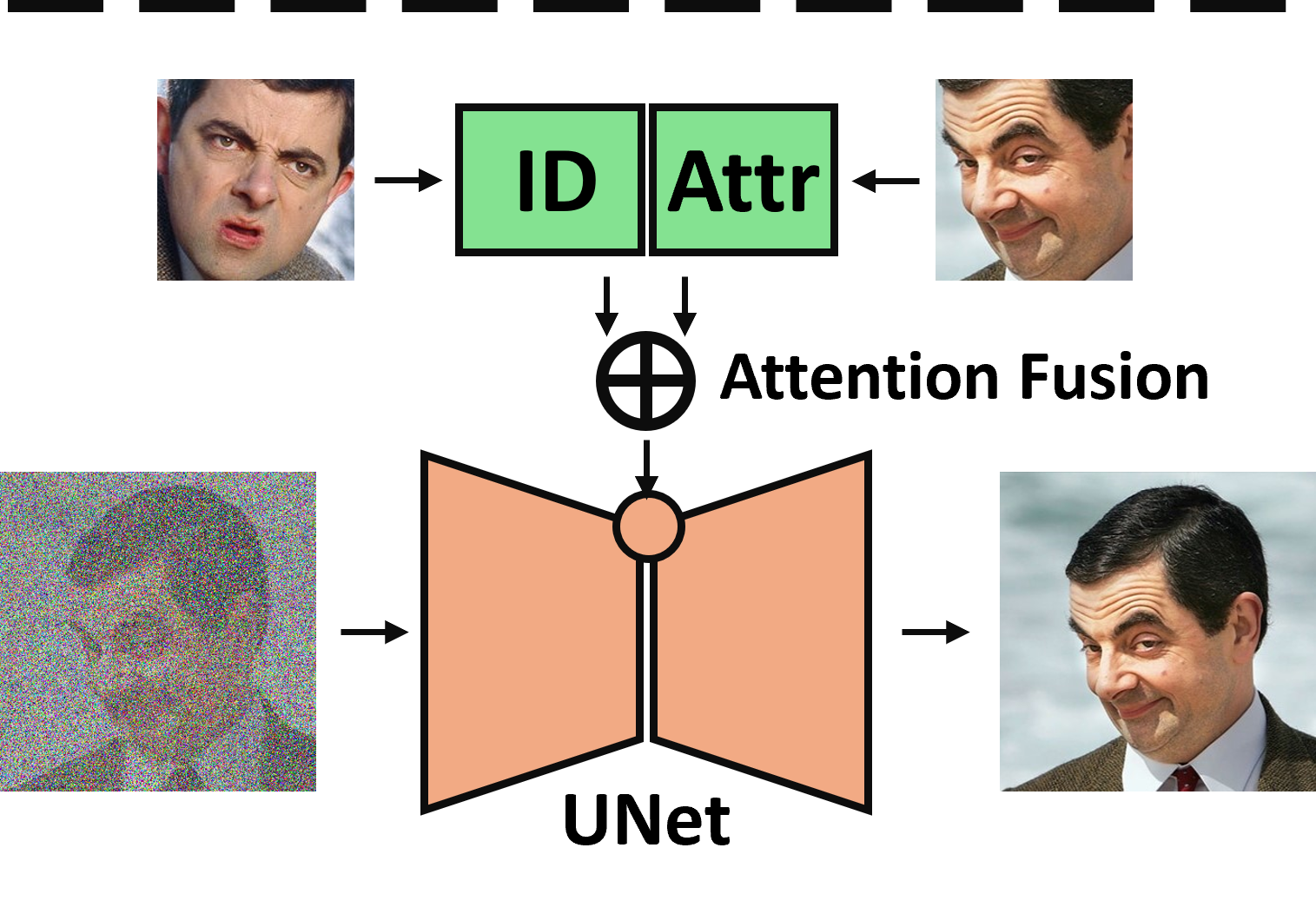}
    \caption{Proposed decoupled facial condition injection.}
    \label{fig:decouple-injection}
    \end{subfigure}

    \caption{Various designs for condition injection. Visual embeddings are injected into the middle attention layers of the denoising UNet in LDMs, with cross-attention adapters~\cite{ye2023ipadapter,li2023gligen}.}
    \label{fig:two-injections}
\end{figure}

\subsection{Decoupled facial condition injection}

In general-purpose text-to-image generation, content control relies on text prompts, but their coarse granularity makes fine-grained facial conditioning challenging. A common approach to address this issue is injecting facial features into the cross-attention layers of the denoising UNet as a task-specific condition (Figure~\ref{fig:two-injections}). 
Given the need for identity-constrained facial conditioning with decoupled identity and attribute conditioning, as discussed in the introduction, we introduce a decoupled facial condition injection strategy (Figure~\ref{fig:decouple-injection}) with two key designs: data-wise condition decoupling and the two-path injection.

\label{sec:data-prepare}

Previous methods often use identical images with augmentation to create sample-condition pairs when training diffusion models (Figure~\ref{fig:aug-injection}).
Using identical images for condition generation risks identity and attribute leakage, as both conditions originate from the same image. Standard augmentations (\eg, resize-crop, color jitter, horizontal flipping) are insufficient to fully differentiate attribute features, making it challenging to achieve decoupled conditions.

To address this issue, we employ facial image pairs from the same identity but with different attributes. This pairing explicitly decouples identity and attribute features, as identity features from the same individual are assumed to be similar, while attributes vary across images.

\label{sec:feature-design}

We adopt a two-path design that separates identity and attribute features, enabling independent processing of each feature type (Figure~\ref{fig:decouple-injection}). After independent computation, the identity and attribute features are fused with an attention fusion module. The fused features are then fed into a GLIGEN injection adapter~\cite{li2023gligen}, which injects facial condition into the UNet via a cross-attention mechanism.

The identity features of the source face are computed using a pretrained ArcFace~\cite{deng2019arcface} recognizer and serve as the main identity representation. 
Leveraging prior knowledge from upstream facial tasks, identity features capture a dense representation of the source identity. To enrich this representation, we extend the $d$-dimension feature to $n\times d$ dimensions using an MLP, and chunk them into a $n$-token sequence $c_{\mathrm{face}}\in\mathbb{R}^{n\times d}$.
To enhance spatial details, we introduce a visual embedding calculated from DINOv2~\cite{oquab2023dinov2}, which captures fine-grained visual information such as skin texture, enhancing the identity representation. 
We use ArcFace features $c_{\mathrm{face}}$ as the query and spatial features $c_{\mathrm{dino}}$ as the key-value sequence,
and formulate the identity features:
\begin{equation}
    c_{\mathrm{id}} = c_{\mathrm{face}} + \lambda_{\mathrm{id}}\mathrm{Attention(c_{\mathrm{face}}, c_{\mathrm{dino}}, c_{\mathrm{dino}})},
    \label{eq:id-fusion}
\end{equation}
where $\lambda_{\mathrm{id}}$ is a spatial enhancement factor, determining the extent to which facial features are enhanced.

Attribute features $c_{\mathrm{attr}}$ are calculated from the target image, capturing its face attributes. To achieve this, we input the target face to the 3-layer downsampling network of SimSwap~\cite{chen2020simswap} to obtain the expression features. In SimSwap the obtained attribute features provide superior representation to achieve a state-of-the-art attribute consistency. Inspired by this, we employ these features as the attribute condition. In addition to expression and pose, this latent-space representation also offers rich spatial semantics.

To achieve identity-constrained conditioning, we fuse the identity and attribute features by querying the identity embedding $c_{\mathrm{id}}$ to the attribute $c_{\mathrm{attr}}$, using an attention layer: 
\begin{equation}
    c_{\mathrm{fuse}} = c_{\mathrm{id}} + \lambda_{\mathrm{fuse}}\mathrm{Attention(c_{\mathrm{id}}, c_{\mathrm{attr}}, c_{\mathrm{attr}})}.
    \label{eq:attr-fusion}
\end{equation}
The fusion factor $\lambda_{\mathrm{fuse}}$ controls the balance between identity and attribute conditioning.
Setting this factor to 0 disables attribute features, resulting in ID-only conditioning.

% Additionally, we estimate the 3D pose of the target face using InsightFace~\cite{insightface}, representing it as a Fourier-based pose embedding similar to the time embedding used in diffusion models~\cite{ho2020ddpm}. This pose embedding is concatenated to the fused feature $c_{\mathrm{fuse}}$. 

\begin{figure*}[t]
    \centering
    \includegraphics[width=0.8\linewidth]{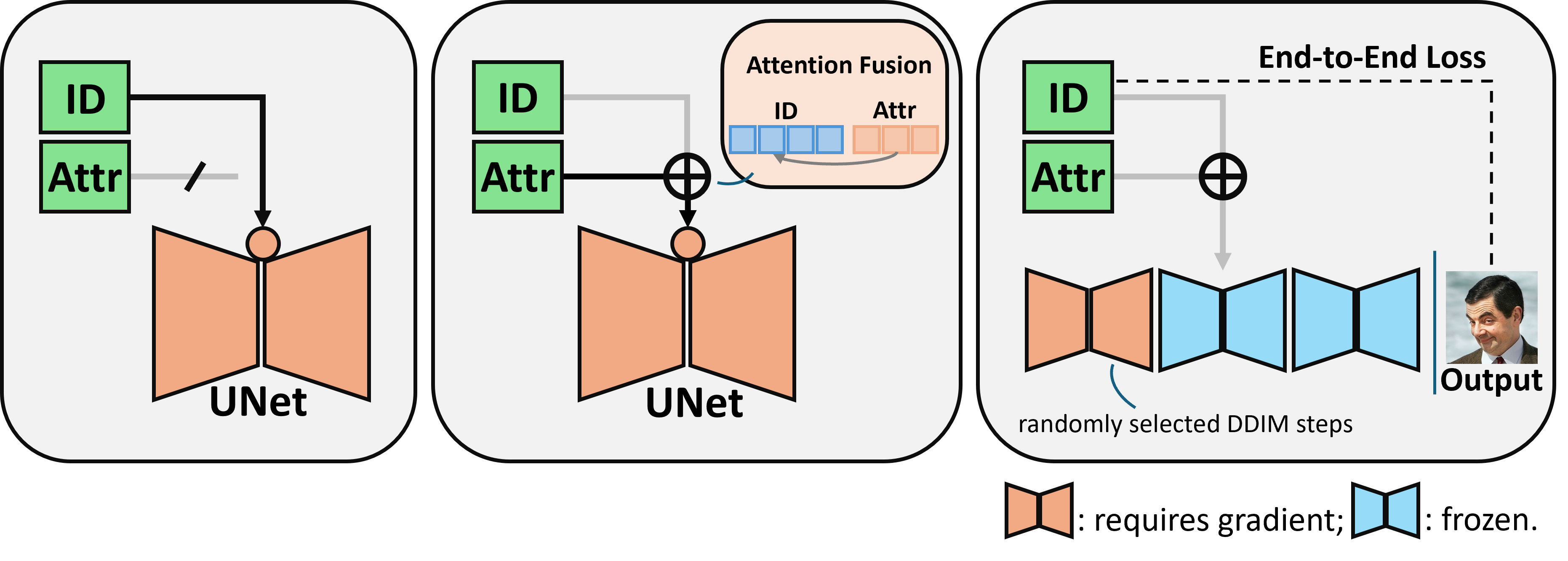}
    \caption{Diagram of the proposed multi-stage training scheme with identity-constrained facial conditioning.}
    \label{fig:diagram-phase1}
\end{figure*}

\subsection{Identity-constrained facial conditioning}

We propose a multi-stage identity-constrained facial conditioning scheme for face swapping, prioritizing the preservation of the source identity while enabling controlled adaptation of target attributes, such as expressions and poses. Figure~\ref{fig:diagram-phase1} illustrates the complete training pipeline, which begins with identity conditioning, progresses to attribute alignment, and concludes with an end-to-end refinement. The following sections introduce this training paradigm.

\subsubsection{Stage 1: identity-oriented tuning}

In the initial stage, the training objectives are:
\begin{enumerate}
    \item adapting the pretrained text-to-image generative model to face swapping task, with conditional foreground inpainting and background preservation;
    \item narrowing the solution space to outputs consistent with the source identity, minimizing deviations. 
\end{enumerate}
The first objective can be achieved with a standard inpainting training process. We extend the linear input layer of the UNet $\epsilon_\theta(\cdot)$ to accept an augmented input, including the noisy latent feature $x_t$, the inpainting region mask $m$, and the background context $(1-m)\odot x_t$. Let $\mathcal{C}$ denote all the conditions injected into the model, such as time embedding and language tokens. The predicted noise at time step $t$ is:
\begin{equation}
    \hat{\epsilon}_t = \epsilon_\theta\left(x_t, m, (1-m)\odot x_t; \ \mathcal{C}\right).
\end{equation}
We can incorporate this adapted noise prediction into eq.~\ref{eq:ori-kld-diff-loss} and perform the standard diffusion training.  Since the model is aware of the foreground and background context, it will soon learn to copy the background region and only generate the foreground guided by the conditions $\mathcal{C}$. During both training and inference, we obtain the mask $m$ with an off-the-shelf face detector.

To achieve the remaining objective, we use exclusively identity conditions without attribute constraints. \ie, we inject $c_\mathrm{fuse}$ into the model with $\lambda_\mathrm{fuse}$ set to 0.
This selective conditioning reduces the solution space to identity-consistent outputs, as shown in Figure~\ref{fig:solution-region}.

\subsubsection{Stage 2: identity-constrained attribute tuning for expression and pose alignment}

Upon establishing identity preservation, the model proceeds to a second stage in which attribute conditions are appended to the condition set $\mathcal{C}$.
This is done by injecting the fused embedding $c_{\mathrm{fuse}}$ with $\lambda_\mathrm{fuse}$ set to one.
This fused condition guides the model to adapt to these target attributes while maintaining the identity constraint established in Stage 1.

Two concerns should be addressed during this training period. First, the attribute condition injection should carefully avoid catastrophic forgetting of the injected identity features. We adopt zero-initialization at the output linear weights of the fusion module's attention layer (in eq.~\ref{eq:attr-fusion}) to cold start the attribute features.
Second, the already sufficient training of facial generation with the identity condition might become overconfident and fail to incorporate the attribute. This risk is amplified because the attribute is often a coarser condition, as discussed in the introduction. To enforce attribute-aware tuning, we weaken the control strength of the identity condition during this stage by decreasing the spatial enhancement factor $\lambda_{\mathrm{id}}$ in eq.~\ref{eq:id-fusion} to 0.2.

\subsubsection{Stage 3: Post-training identity refinement}

\begin{figure*}[t]
    \centering
    \includegraphics[width=0.9981\linewidth]{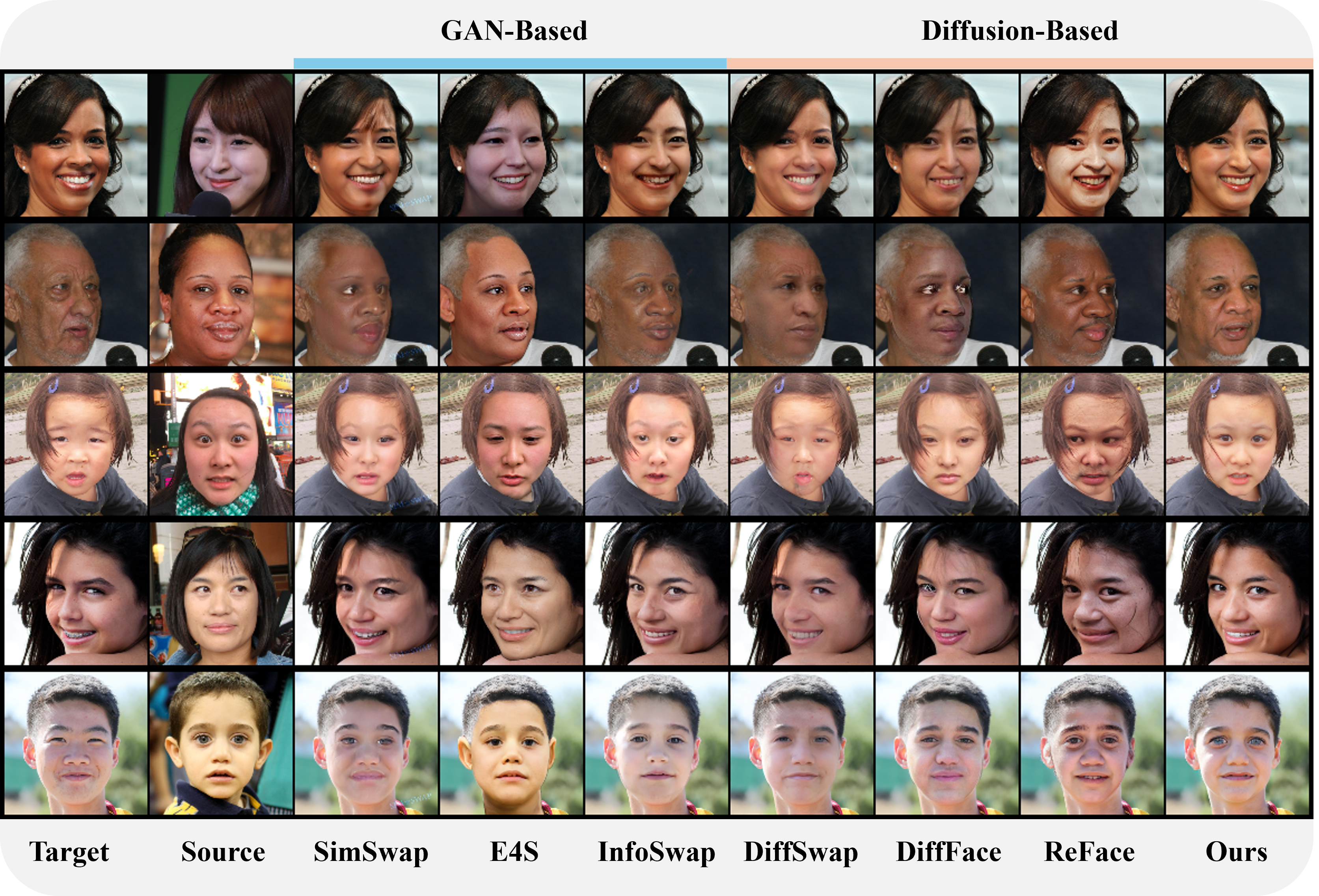}
    \caption{Face-swapping results using various approaches, evaluated on FFHQ.}
    \label{fig:compare-ffhq}
\end{figure*}

In the previous stages, we train a face-swapping diffusion model with both identity and attribute awareness. All training follows the standard diffusion process described in Section~\ref{sec:preliminary-of-diff-model}, with noise prediction in eq.~\ref{eq:ori-kld-diff-loss}. Notably, this conditional training is implicit, lacking explicit supervision of identity similarity. In contrast, most GAN-based approaches use explicit
identity similarity loss (ID loss)~\cite{chen2020simswap,li2019faceshifter,nirkin2019fsgan}. To ensure identity similarity, DiffSwap~\cite{zhao2023diffswap} and DiffFace~\cite{kim2022diffface} directly add this term to the standard diffusion noise-prediction loss (eq.~\ref{eq:ori-kld-diff-loss}). However, this noise-prediction loss is theoretically the ELBO, and adding terms to it relaxes the bound, resulting in decreased sample quality.  REFace~\cite{baliah2024reface} applies this ID loss to the final sampled image, accompanied by standard diffusion noise-prediction training. This approach is limited by computationally intensive multi-step DDIM sampling.

We propose to separate the identity supervision into a distinct training phase. After condition injection, we treat the entire 50-step DDIM sampling as a cascaded end-to-end generative model. Thus, we can utilize end-to-end supervision, such as identity loss and GAN losses, on this pipeline. Inspired by previous studies on DDIM tuning and the deep reward model~\cite{wu2024deepreward}, because of the massive intermediate memory cost for back-propagation, we randomly sample $k$ steps from the entire 50 DDIM steps during each mini-batch training, and we calculate gradient only on those $k$ steps.
We use a combined loss function for this refinement with two partitions: GAN losses $\mathcal{L}_\mathrm{adv}$ and ID loss $\mathcal{L}_\mathrm{id}$:
\begin{equation}
    \mathcal{L} = 
    \lambda_\text{adv} \mathcal{L}_\text{adv} +
    \lambda_\text{id} \mathcal{L}_\text{id}.
\end{equation} 
We use an SNGAN~\cite{miyato2018sngan} discriminator and use hinge losses.

\section{Experiments}

\textbf{General settings.} We employ Stable Diffusion 1.5~\cite{rombach2022latentdiff} as the pretrained foundation model.
We collect 4.5 million images paired with identities from the Internet. Most are extracted from videos where the same character performs different actions in a facial pair. Once data collection is complete, we use the BLIP-2~\cite{li2023blip2} model to annotate each image, enriching the dataset with descriptive text captions for text-to-image model input.
In stage-1 and stage-2 training, we use this paired dataset to train the models. In the third stage, we perform post-training using randomly paired facial data filtered from LAION-5B~\cite{schuhmann2022laion5b}.

We evaluate the results using FFHQ~\cite{karras2019stylegan-ffhq}.
We refer to the REFace evalation protocol~\cite{baliah2024reface}, using the 1000 pairs of images from the validation split as the test dataset.
In all evaluations, we use an output size of $512\times 512$ pixels. 
We select several cutting-edge methods to construct a strong baseline. For GAN-based approaches, we test SimSwap~\cite{chen2020simswap}, InfoSwap~\cite{gao2021infoswap}, and E4S~\cite{liu2023e4s}. Additionally, we evaluate DiffFace~\cite{kim2022diffface}, DiffSwap~\cite{zhao2023diffswap}, and REFace~\cite{baliah2024reface} which are all based on diffusion models.

\begin{figure}
    \centering
    \includegraphics[width=0.9981\linewidth]{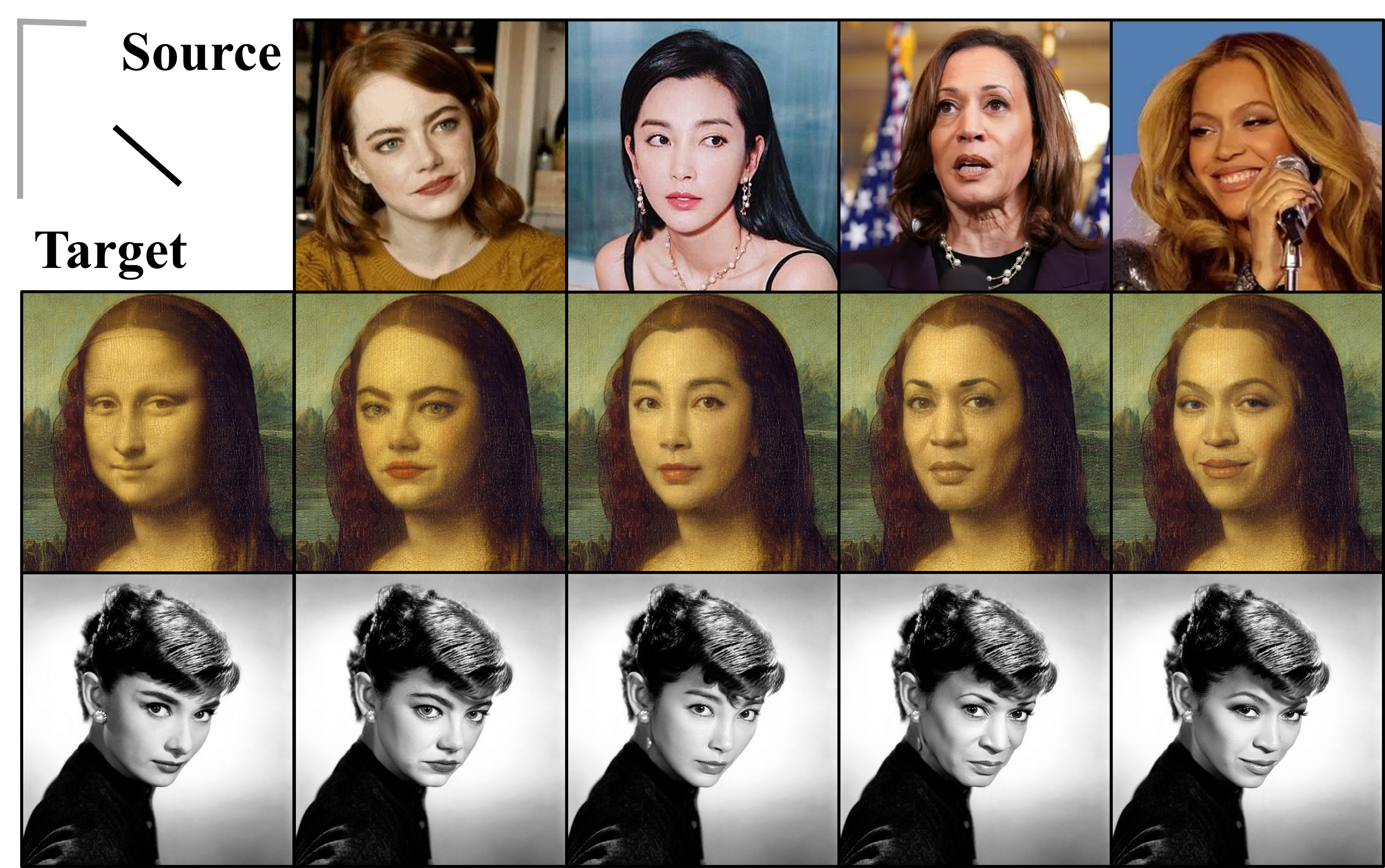}
    \caption{Stylized face swapping with our model.}
    \label{fig:stylization}
\end{figure}

\begin{figure}
    \centering
    \includegraphics[width=0.9981\linewidth]{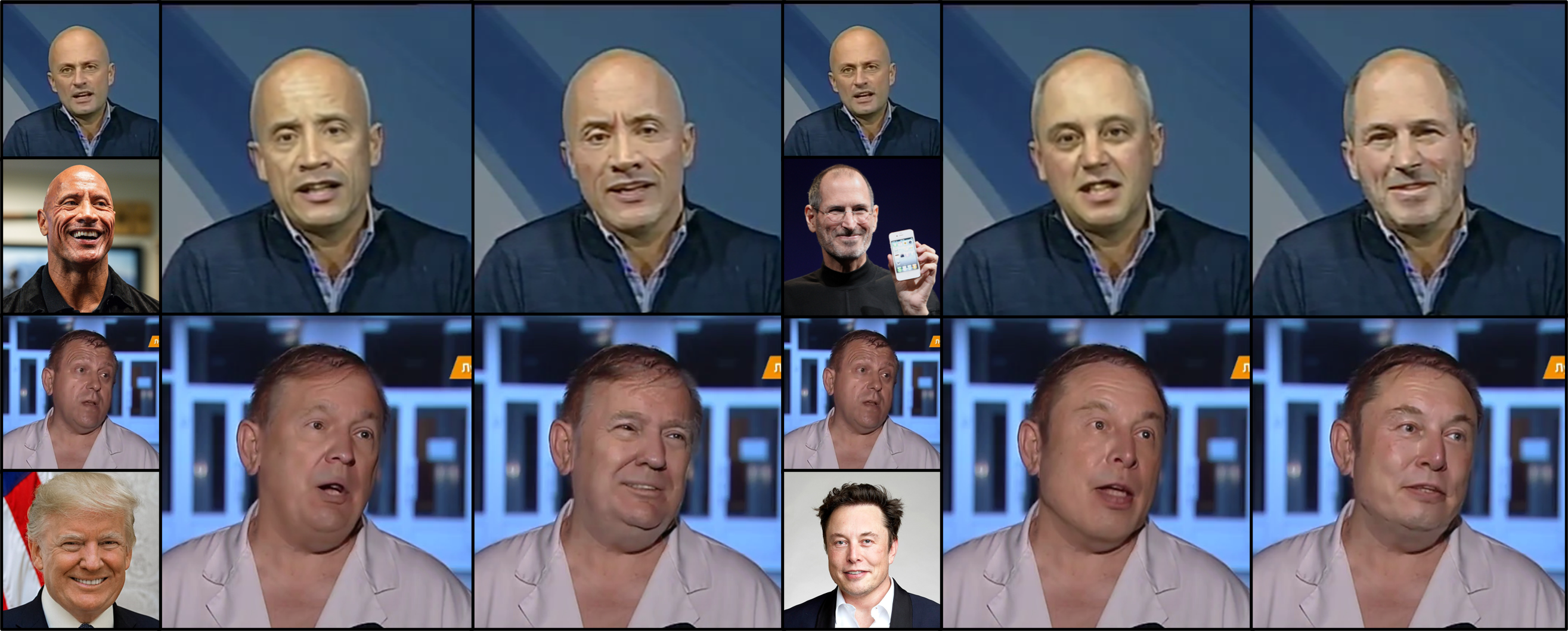}
    \caption{Comparison of non-decoupled (left per group) and decoupled (right per group) condition injection.}
    \label{fig:comp-injection}
\end{figure}

\subsection{Qualitative results}

\textbf{Comparison on FFHQ.} Figure~\ref{fig:compare-ffhq} shows face-swapping results on FFHQ of different approaches. Compared with existing GAN-based and diffusion-based approaches, our proposed model generates faces with better identity preservation and higher fidelity. Our results demonstrate strong attribute consistency, avoiding unrealistically distorted faces.

Compared to ours, existing diffusion approaches struggle to adhere to identity and attribute constraints. In several cases, they generate low-quality results with significant artifacts. We emphasize that identity-constrained training and decoupled condition injection are essential to avoid suboptimal diffusion training and dual-condition conflicts.

In addition to superior identity and attribute consistency, our model generates more natural skin texture and lighting conditions compared to GAN-based approaches. This demonstrates the advantage of diffusion-based approaches, utilizing the strong generative ability of pretrained foundation models. Unlike task-specific models, using general-purpose foundation models with inherent generalization helps handle corner cases and improve generative robustness. This further demonstrates the superiority of diffusion-based face-swapping methods.

\paragraph{Stylized face swapping.} We illustrate the face-swapping results using stylized targets in Figure~\ref{fig:stylization} and Figure~\ref{fig:teaser} to demonstrate the out-of-distribution robustness of our approach. When swapping faces with paintings and stylized photos, our model effectively maintains the texture style, and the generalized faces harmoniously blend into the background.  Since images with significant style are rare in our training dataset (approximately less than 1\%), these results suggest that our model has exceptional generalization against out-of-distribution data. This is unsurprising, as we use pre-trained foundation diffusion models that have already been trained on large-scale datasets, including stylized images. Most notably, this is a unique advantage of diffusion-based approaches: the foundation models provide off-the-shelf generalized generation capability, allowing us to focus solely on facial condition injection.
\newcommand{\stagewisesubfigwid}{0.9981\linewidth}
\begin{figure}
    \centering        \includegraphics[width=0.9999\linewidth]{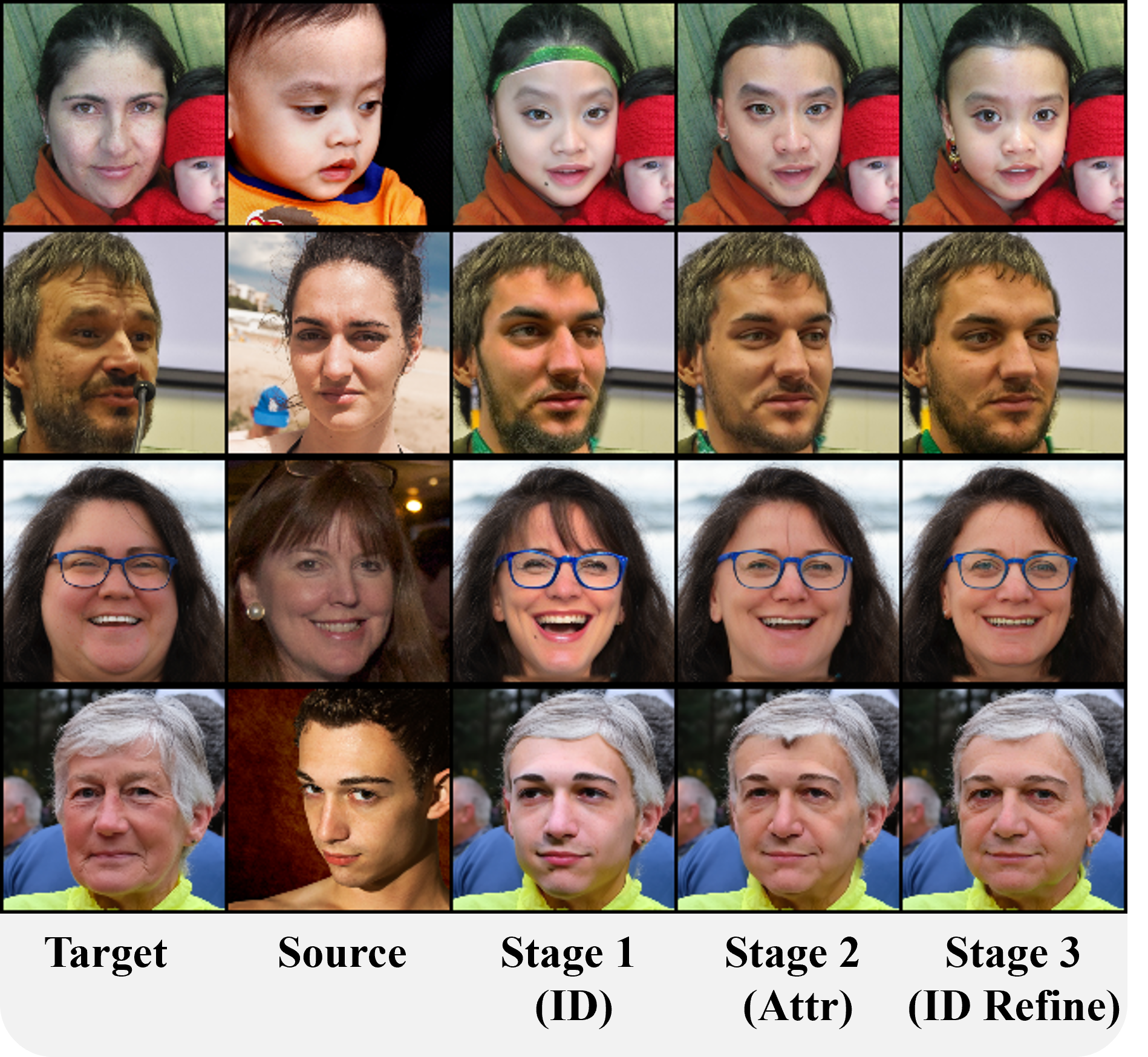}
    \caption{Results of our model after each training stage.}
    \label{fig:stage-wise}
\end{figure}

\paragraph{Comparing non-decoupled and decoupled injection.} Figure~\ref{fig:comp-injection} shows face-swapping results without and with decoupled injection. The non-decoupled results were obtained by using augmented identical image conditioning and merging the first two training stages with the same training volume. We observe that without decoupled injection, the model overfits to target attribute following and fails to preserve identity. This demonstrates the effectiveness of our proposed decoupled condition injection.

\paragraph{Effectiveness of multi-stage training.} We evaluate the results of different training stages to demonstrate the effectiveness of the proposed multi-stage training scheme. See Figure~\ref{fig:stage-wise}. In all stage-1 results, the identity-constrained face-swapping is reasonable but fails to capture the target face's expression. After attribute tuning in the second stage, the model shows improvement, with outputs better following the attributes. Surprisingly, in addition to expression, pose, and gaze, we observe improvements in lighting conditions in several cases. This unexpected bonus is likely due to the use of a pretrained SimSwap attribute encoder, which also captures the lighting conditions of the target image. We also observe a face shape adjustment after this stage, where the model aligns the target expression with the source face shape. The investigation of lighting and face shaping in face swapping is preliminary. We believe our unified pipeline is likely to inspire further studies on finer-grained attribute control. After the final training stage, identity similarity and generation fidelity are significantly enhanced in most cases, proving the necessity of end-to-end post-training.

\subsection{Quantatative results}

\begin{table}[t]
    \centering
    \scalebox{0.8}{
    \begin{tabular}{lrrrrr}
         \toprule
         &&&& \multicolumn{2}{c}{\textbf{ID Retrieval$\uparrow$}}  \\
         \textbf{Method} &  \textbf{FID$\downarrow$} & \textbf{Pose$\downarrow$} & \textbf{Expr.$\downarrow$} & \textbf{Top-1} & \textbf{Top-5} \\
         \midrule
         \textit{\textbf{GAN-based:}} \\
         SimSwap~\cite{chen2020simswap} & 13.74 & \underline{2.62} & \textbf{0.95} &\underline{93.37}&97.29\\
         
         E4S~\cite{liu2023e4s} &12.22&4.55&1.32 
 &77.80&87.40\\
        InfoSwap~\cite{gao2021infoswap} & \underline{4.26} & 3.26 & 1.00 & 92.82 & \underline{97.69} \\
         \multicolumn{2}{l}{\textit{\textbf{Diffusion-based:}}} \\
         DiffFace~\cite{kim2022diffface} & 8.82 & 3.76 & 1.31 & 91.50 &  97.50\\
         DiffSwap~\cite{zhao2023diffswap} & 5.80 & \textbf{2.43} & 1.01 & 67.00 & 81.90  \\
         REFace~\cite{baliah2024reface} & 5.62 & 3.75 & 1.04 & 89.10 & 96.10\\
         Ours & \textbf{3.61} & 3.69 & \underline{0.97} & \textbf{97.90} & \textbf{99.70} \\
         \bottomrule
    \end{tabular}
    }
    \caption{Metrics on FFHQ compared with previous approaches.}
    \label{tab:comparision-ffhq}
\end{table}

\begin{table}[t]
    \centering
    \scalebox{0.8}{
    \begin{tabular}{llrlrlr}
           \toprule
         \textbf{\#} & \textbf{ID} & {(\%)} & \textbf{Attr.} & {(\%)} & \textbf{Fidelity} & {(\%)} \\
         % \cmidrule{2-3}  \cmidrule{4-5}  \cmidrule{6-7} 
         \midrule
         1 & E4S & 36.7 & \underline{\textbf{Ours}} & 33.2 &\underline{\textbf{Ours}} & 57.1  \\
         2 & \underline{\textbf{Ours}} & 31.6 & D.Face & 30.6 & D.Swap & 15.3 \\
         3 & REFace & 10.2 & S.Swap & 13.8 & D.Face & 14.3 \\
         4 & D.Swap & 8.7 & D.Swap & 10.2 & S.Swap & 6.6 \\
         \bottomrule
    \end{tabular}
    }
    \caption{Human evaluation results. We ask participantsto vote for their preferred outputs from all evaluated methods according to three criteria. We present the top four methods for each evaluation with their respective proportions.}
    \label{tab:user-study}
\end{table}

\paragraph{Comparison on FFHQ.} See Table~\ref{tab:comparision-ffhq}. We compare our method with previous GAN-based and diffusion-based approaches. Following prior work, we evaluate four quantitative metrics: Fr\'echet Inception Distance (FID), pose distance, expression, and identity retrieval accuracy. To measure the pose distance, we estimate the 3D roll-yaw-pitch pose vector using HopeNet~\cite{doosti2020hopenet}. And we estimate expression landmarks with Deep3DFaceRecon~\cite{deng2019deep3dfacerecon}. For ID retrieval, we first calculate the face feature vector with ArcFace~\cite{deng2019arcface}, then calculate the similarity between the output face and all source faces in the dataset. We then select the sources with the highest similarities to the output and calculate accuracy.

As a result, our proposed model achieves outstanding metrics. First, our FID score is the most competitive, significantly improving on existing approaches. As FID is considered a sensitive indicator of overall sample quality, this improvement implies strong generative quality in our method. This is demonstrated by the visual results in Figure~\ref{fig:compare-ffhq}, as discussed above. Notably, this FID superiority is significant even compared to other diffusion-based approaches. We attribute this to two factors. First, we address the condition completion issue, which helps overcome the sub-optimal training. Second, we adopt a unique ID loss training without appending it to noise-prediction loss, thus relaxing the ELBO. In contrast, DiffFace~\cite{kim2022diffface} and DiffSwap~\cite{zhao2023diffswap} use an additively combined loss function, which may reduce generation quality.
In addition to exceptional FID performance, our model achieves the best ID retrieval accuracy, indicating the success of identity-constrained training. Although our model does not surpass others in pose and expression measurements, it remains competitive. Notably, our expression score on FFHQ matches the best score achieved by SimSwap~\cite{chen2020simswap}.

\vspace{-0.4cm}\paragraph{Human evaluation.} See Table~\ref{tab:user-study}. Since each quantitative metric has its own bias and may not align with human perception~\cite{zhang2018lpips}, we conducted a human evaluation to better compare the performance of various methods. The evaluation was conducted on a diverse group considering gender, age, and culture. 39 participants voted on FFHQ face-swapping results based on their preferences for identity similarity, attribute consistency, and fidelity. respectively. We summarize the evaluation results in Table~\ref{tab:user-study}. In this study, our model performs remarkably, especially with over 50\% voting on the fidelity. Our method also achieves the best attribute consistency with 33.2\% of votes and the second-best identity preservation with 31.6\% of votes. Surprisingly, the E4S method, with the most votes for identity, performs poorly in the other two evaluations. As shown in Figure~\ref{fig:compare-ffhq}, this approach may generate artifacts in some cases. It may overfit to identity preservation. In contrast, our approach shows overall superiority in all three subjective evaluations, demonstrating a good balance between facial condition adherence and generation quality.

\section{Conclusion}
In this paper, we enhance diffusion-based face swapping by carefully decoupling two types of facial conditioning: identity and attribute. We train the diffusion model using an identity-constrained scheme with attribute tuning to address identity-attribute imbalance introduced by condition completion. Experimental results demonstrate that our approach significantly improves face swapping, effectively preserving both input identity and attributes.

Notably, face swapping is a specific instance of general conditional generation. Identity preservation and customization~\cite{ruiz2023dreambooth, ye2023ipadapter,li2023gligen} are crucial in these tasks. Our decoupled condition injection and identity-constrained approach could serve as a foundation for studies on general tasks, as it addresses a particular instance of multi-condition injection. 
% Intuitively, these interactions among conditions exist in broader image generation, requiring further exploration. 
We intend to delve deeper into this line of methods, generalizing them to more applications.

\section*{Acknowledgements}
This study was supported in part by National Key R\&D Program of China Project 2022ZD0161100, in part by the Centre for Perceptual and Interactive Intelligence, a CUHK-led InnoCentre under the InnoHK initiative of the Innovation and Technology Commission of the Hong Kong Special Administrative Region Government, in part by NSFC-RGC Project N\_CUHK498/24, and in part by Guangdong Basic and Applied Basic Research Foundation (No.2023B1515130008, XW).
{
    \small
    \bibliographystyle{ieeenat_fullname}
    \bibliography{main}
}

% WARNING: do not forget to delete the supplementary pages from your submission 
% \input{sec/X_suppl}

\end{document}